\documentclass{article}

\usepackage{microtype}
\usepackage{graphicx}
\usepackage{subcaption}
\usepackage{booktabs} 

\usepackage{hyperref}

\usepackage[preprint]{icml2026}

\usepackage{amsmath}
\usepackage{amssymb}
\usepackage{mathtools}
\usepackage{amsthm}
\usepackage{array}

\usepackage{multicol}
\usepackage{multirow}
\usepackage{xcolor}
\usepackage{enumitem}

\usepackage[capitalize,noabbrev]{cleveref}
\usepackage{siunitx}
\usepackage{booktabs}
\usepackage{dcolumn}
\usepackage[normalem]{ulem} 
\newcolumntype{d}{D{.}{.}{2}}
\theoremstyle{plain}

\theoremstyle{definition}

\theoremstyle{remark}

\icmltitlerunning{EarthSpatialBench: Benchmarking Spatial Reasoning Capabilities of Multimodal LLMs on Earth Imagery}

\begin{document}

\twocolumn[
  \icmltitle{EarthSpatialBench: Benchmarking Spatial Reasoning Capabilities of Multimodal LLMs on Earth Imagery}
  \icmlsetsymbol{equal}{*}

  \begin{icmlauthorlist}
    \icmlauthor{Zelin Xu}{uf}
    \icmlauthor{Yupu Zhang}{uf}
    \icmlauthor{Saugat Adhikari}{iu}
    \icmlauthor{Saiful Islam}{uf}
    \icmlauthor{Tingsong Xiao}{uf}
    \icmlauthor{Zibo Liu}{uf}
    \icmlauthor{Shigang Chen}{uf}
    \icmlauthor{Da Yan}{iu}
    \icmlauthor{Zhe Jiang}{uf}
  \end{icmlauthorlist}

  \icmlaffiliation{uf}{Department of Computer \& Information Science \& Engineering, University of Florida, Gainesville, FL, USA}
  \icmlaffiliation{iu}{Department of Computer Science, Indiana University Bloomington, Bloomington, IN, USA}

  \icmlcorrespondingauthor{Zhe Jiang}{zhe.jiang@ufl.edu}
  \icmlkeywords{Machine Learning, ICML}

  \vskip 0.3in
]
\printAffiliationsAndNotice{}  

\begin{abstract}
Benchmarking spatial reasoning in multimodal large language models (MLLMs) has attracted growing interest in computer vision due to its importance for embodied AI and other agentic systems that require precise interaction with the physical world.  
However, spatial reasoning on Earth imagery has lagged behind, as it uniquely involves grounding objects in georeferenced images and quantitatively reasoning about distances, directions, and topological relations using both visual cues and vector geometry coordinates (e.g., 2D bounding boxes, polylines, and polygons).
Existing benchmarks for Earth imagery primarily focus on 2D spatial grounding, image captioning, and coarse spatial relations (e.g., simple directional or proximity cues). They lack support for quantitative direction and distance reasoning, systematic topological relations, and complex object geometries beyond bounding boxes.
To fill this gap, we propose \textbf{EarthSpatialBench}, a comprehensive benchmark for evaluating spatial reasoning in MLLMs on Earth imagery. The benchmark contains over 325K question–answer pairs spanning: (1) qualitative and quantitative reasoning about spatial distance and direction; (2) systematic topological relations; (3) single-object queries, object-pair queries, and compositional aggregate group queries; and (4) object references expressed via textual descriptions, visual overlays, and explicit geometry coordinates, including 2D bounding boxes, polylines, and polygons. We conducted extensive experiments on both open-source and proprietary models to identify limitations in the spatial reasoning of MLLMs.
\end{abstract}    
\section{Introduction}
\label{sec:intro}

Benchmarking the spatial reasoning capabilities of multimodal large language models (MLLMs)~\cite{achiam2023gpt, liu2023visual, bai2023qwen, chen2024internvl} has attracted growing interest in computer vision~\citep{yang2025thinking, deng2025internspatial, cheng2024spatialrgpt, Chen_2024_CVPR, Ma_2025_ICCV}. Spatial reasoning involves analyzing and interpreting the positions, orientations, and spatial relationships of objects based on 2D photos. It is crucial for embodied AI and other agentic systems that require precise interaction with the physical world.  A large number of benchmarking datasets have been developed for natural images~\cite{deng2025internspatial,Chen_2024_CVPR,cheng2024spatialrgpt,shiri-etal-2024-empirical,cai2025spatialbot,zhang2025open3d,li2024cambrian1}.

However, benchmarking spatial reasoning on Earth imagery has lagged behind. Earth imagery, spanning satellite, aerial, and drone observations, is pervasive. Benchmarking and developing spatial reasoning capabilities of MLLMs on Earth imagery has the potential to address many grand societal challenges, such as natural disaster response, urban planning, precision agriculture, and ecological monitoring~\citep{stallings2003methods, shi2020disaster, shekhar2015spatial, shekhar2007encyclopedia}.  
For example, during a deadly flood disaster, MLLMs can assist first responders in quickly locating the impacted communities, estimating the economic loss, and planning rescue operations (e.g., counting damaged houses within certain neighborhood blocks, and locating them within a distance range from major roads).

\begin{table*}\scriptsize
\centering
\caption{A comparison of existing benchmark datasets for MLLM on Earth imagery. ($\checkmark$ "Fully Benchmarked"; $\triangle$ "Partially Benchmarked"; $\times$ "Not Benchmarked")}
\begin{tabular}{l c c ccc}
\toprule

\multirow{2}{*}{\textbf{Datasets}} &
\multirow{2}{*}{\textbf{Number of QA Pairs}} &
\multirow{2}{*}{\textbf{Geometry Type}} &
\multicolumn{3}{c}{\textbf{Spatial Reasoning}} \\

\cmidrule(lr){4-6}
& & & \textbf{Distance} & \textbf{Direction} & \textbf{Topological} \\

\midrule

EarthVQA~\citep{wang2024earthvqa} & 209K & BBox & $\times$ & $\times$ & $\times$ \\
GeoChatSet~\citep{kuckreja2024geochat} & 318K & BBox & $\times$ & $\times$ & $\times$ \\
VRSBench~\citep{li2024vrsbench} & 123K & BBox & $\times$ & $\times$ & $\times$ \\
SkyEyeGPT~\citep{zhan2025skyeyegpt} & 968K & BBox & $\times$ & $\times$ & $\times$ \\
XLRS-Bench~\citep{wang2025xlrs} & 46K & BBox & $\times$ & $\triangle$ & $\times$ \\
GeoBench-VLM~\citep{danish2025geobenchvlm} & 10K & BBox & $\times$ & $\triangle$ & $\times$ \\
DisasterM3~\citep{wang2025disasterm3} & 123K & BBox & $\times$ & $\triangle$ & $\times$ \\

\midrule

\textbf{EarthSpatialBench} & \textbf{325K} & \textbf{BBox, Polygon, Polyline}  
& $\checkmark$ & $\checkmark$ & $\checkmark$ \\

\bottomrule
\end{tabular}

\label{tab:dataset_comparison_updated}
\end{table*}

Spatial reasoning on Earth imagery poses several unique challenges. First, compared with natural imagery (on which existing MLLMs are largely pretrained), Earth imagery is often noisy with denser objects (e.g., tens of houses) in a bird's-eye view (e.g., houses are tiny with only roofs visible). Second, an object in Earth imagery can be referred to in multiple ways, including both textual descriptions of visual cues (e.g., object types and spatial context) and vector geometric coordinates (e.g., 2D bounding boxes (bboxes), polylines, and polygons). Third, spatial reasoning involves the estimation of quantitative distance and direction (azimuth angle), as well as complex topological relations between objects in rich geometries (e.g., counting houses within a certain distance of a river, locating houses within a park polygon). 

Table~\ref{tab:dataset_comparison_updated} summarizes the existing MLLM benchmarking datasets on Earth imagery. These datasets mainly focus on spatial grounding (e.g., locating 2D bounding boxes of objects on Earth imagery). A few datasets~\cite{wang2025xlrs,danish2025geobenchvlm,wang2025disasterm3} include coarse direction queries (e.g., left of, above, beside), and sometimes the reference to input objects is based on drawing boxes on the original Earth imagery. Almost none of them include quantitative direction (azimuth angle), distance metrics, and spatial topological relations. In addition, these datasets do not incorporate rich object geometries (polyline, polygon). 

To fill this gap, we propose \textbf{EarthSpatialBench\footnote{The complete dataset will be released upon acceptance.}}, a comprehensive benchmark for evaluating spatial reasoning in MLLMs on Earth imagery. The benchmark contains over 325K question–answer pairs spanning: (1) qualitative and quantitative reasoning about spatial distance and direction; (2) systematic topological relations; (3) single-object queries, object-pair queries, and compositional aggregate group queries based on distance, direction, or topological relation; and (4) object references expressed via textual descriptions, visual overlays, and explicit geometry coordinates, including 2D bounding boxes, polylines, and polygons. We conducted evaluations of multiple MLLMs on our datasets to identify limitations in spatial reasoning capabilities.
\section{Relevant Works}

\subsection{Spatial Reasoning Benchmarks for Natural Images}
A growing number of benchmarking datasets have been developed for MLLM spatial reasoning on natural images. 
InternSpatial~\citep{deng2025internspatial} provides large-scale single- and multi-view spatial reasoning, while SpatialVLM~\citep{Chen_2024_CVPR}, SpatialRGPT-Bench~\citep{cheng2024spatialrgpt}, and Spatial-MM~\citep{shiri-etal-2024-empirical} introduce human-annotated QA for relational understanding. Other datasets—including SpatialBench~\citep{cai2025spatialbot}, Open3D-VQA~\citep{zhang2025open3d}, and CV-Bench~\citep{li2024cambrian1}—focus on object-centric relations such as proximity, counting, or size comparison. However, these benchmarks rarely evaluate continuous metric reasoning, angular direction estimation, or formal topological relations, and they generally lack the polygonal and polyline structures intrinsic to geospatial analysis. EarthSpatialBench complements this line of work by introducing spatial reasoning grounded in real Earth imagery and GIS-style geometric interactions.

\subsection{Multimodal Large Language Models}

Early versions of multimodal large language models (MLLMs), e.g.,~LLaVA~\cite{liu2023visual}, mainly focus on vision–language instruction tuning to expand the modality of LLMs from text-only to multimodal text-image capabilities. Subsequent open-source MLLMs scale model size and data while improving spatial reasoning, OCR, and long-context understanding, such as the Qwen3-VL family, built on the Qwen3-VL framework with dense and MoE variants for high-capacity vision–language reasoning~\cite{yang2025qwen3}. In parallel, the InternVL series culminates in InternVL3.5, which employs cascade reinforcement learning and resolution routing to boost multimodal reasoning and efficiency across images, documents, and video~\cite{wang2025internvl35}. In particular, Qwen3-VL claims a major advancement in 2D and 3D grounding and complex spatial reasoning capabilities. Meanwhile, there are commercial closed-source multimodal LLMs, such as OpenAI's GPT-4o and GPT-5~\cite{openai2025gpt5}, and Google's Gemini 2.5 Pro~\cite{deepmind2025gemini25pro}.

\begin{figure*}[ht]
    \centering
    \includegraphics[width=\textwidth]{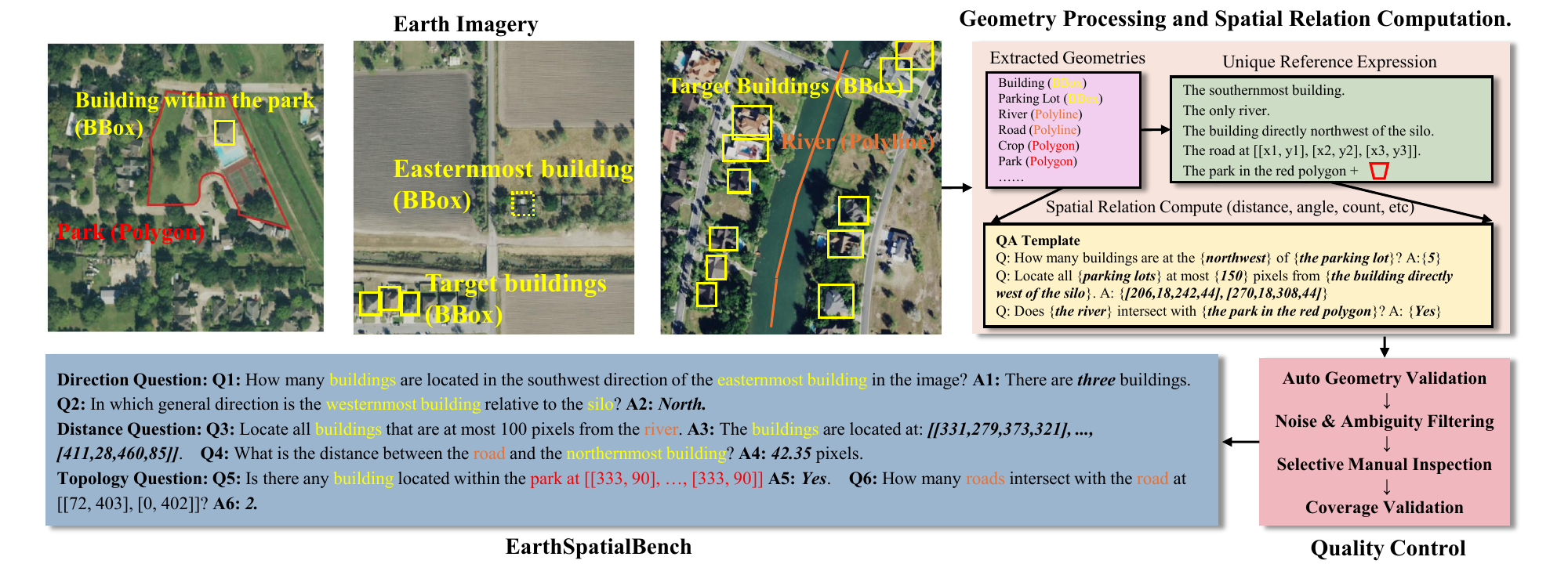}
    \caption{Overview of EarthSpatialBench. The benchmark spans diverse spatial relations evaluated over multiple geometry types, object representation modalities, and question formats, enabling comprehensive assessment of spatial reasoning. High-resolution Earth imagery and geometric annotations are collected from SatlasPretrain. Spatial relations (distance, direction, topology) are computed using standard GIS procedures and combined with curated question templates to generate diverse reasoning tasks. A multi-stage quality control process ensures geometric accuracy, semantic clarity, and reliable ground-truth annotations.}
\label{fig:overview}
\end{figure*}

\begin{table}[ht]\small
\centering
\caption{Overview of EarthSpatialBench.}
\begin{tabular}{p{4cm}p{2.3cm}}
\toprule
\textbf{Property} & \textbf{Value} \\
\midrule
Source dataset & SatlasPretrain~\cite{bastani2023satlaspretrain} \\
Sensor type & Optical RGB \\
Image resolution & 1m \\
Image size & 512 $\times$ 512 \\
Total images & 21K \\
Total QA pairs &  325K\\
Bbox objects & 6 classes \\
Polyline objects & 7 classes \\
Polygon objects & 10 classes \\
Average objects per image & 19 \\
Average bboxes per image & 11 \\
Average polylines per image & 7 \\
Average polygons per image & 2 \\
\bottomrule
\end{tabular}
\label{tab:dataset_overview}
\end{table}

\section{EarthSpatialBench}

To guide the spatial reasoning evaluation of MLLMs on Earth imagery, as shown in Figure~\ref{fig:overview}, EarthSpatialBench is designed around three core research questions:
\begin{itemize}[leftmargin=*,noitemsep,topsep=0pt]
\item \textbf{(RQ1) Spatial Relation Understanding:} How well can MLLMs conduct spatial reasoning based on different types of spatial relations on Earth imagery (e.g., distance, direction, and topological relations)?
\item \textbf{(RQ2) Reasoning with Different Geometric Representation:} When performing spatial reasoning, how effectively do MLLMs handle different reference expressions of objects, including textual description (object types, spatial context), visual overlays, and explicit geometric coordinates of objects? How do they perform on objects in different geometric types, such as 2D bounding boxes (bboxes), polygons, and polylines?
\item \textbf{(RQ3) Task Format and Reasoning Difficulty:} How do MLLMs perform across different question formats, including qualitative questions (e.g., discrete choices, Yes or No) versus quantitative (azimuth degree, Euclidean distance), single object pair (exact distance or direction between two objects) versus composite group aggregate questions (e.g., objects within a certain distance or direction range from another object), and counting versus localization questions. 
\end{itemize}
These research questions allow us to systematically characterize the strengths and limitations of current MLLMs in geospatial reasoning on Earth imagery data. Accordingly, EarthSpatialBench is organized around orthogonal dimensions that collectively capture the breadth of spatial reasoning required for Earth imagery. 

\subsection{Evaluation Dimensions}

\textbf{Spatial Relations.} We categorize spatial relations into three main types: distance, direction, and topological relations. Together, they capture the common spatial relationships involved in geospatial reasoning on Earth imagery.
\begin{itemize}[leftmargin=*,noitemsep,topsep=0pt]
    \item \textbf{Distance.} Distance relations describe how far objects or regions are from one another under a metric. They are central to proximity analysis and buffer-based operations. EarthSpatialBench includes tasks that require estimating the distance between two objects, determining whether two objects fall within a specified distance threshold, and counting or identifying objects within a defined buffer.
    \item \textbf{Direction.} Direction relations encode the relative orientation of objects or regions, expressed either as angular measurements or as coarse directional categories (for example, the eight compass directions). Such reasoning is important for navigation, spatial alignment, and understanding scene layout. EarthSpatialBench includes queries that involve estimating angles between objects, assigning objects to directional bins, and counting or locating objects within particular directional sectors.
    \item \textbf{Topological relation.} Topological relations describe how geometric entities connect, overlap, or are embedded within one another, independent of distance or orientation. We focus on two common cases in geospatial analysis: within (contain) and intersect. These tasks involve determining whether one region lies entirely inside another, identifying intersections between geometries, and locating objects that satisfy these relationships.
\end{itemize}
Together, distance, direction, and topology provide a principled decomposition of spatial reasoning that goes well beyond conventional perception benchmarks.

\textbf{Geometry Types.} Spatial relations in Earth imagery are expressed over geometric entities rather than only over discrete object instances. EarthSpatialBench explicitly models three primary geometric types: bounding boxes, polygons, and polylines, which together cover a broad set of common geospatial elements. We exclude point geometries because they are typically not visible at Earth imagery resolution. Although a bounding box is formally a simple polygon, we treat it as a separate category because it serves as a stable, perceptually grounded representation for small objects (e.g., buildings), while polygons preserve the true shapes of larger region-level entities (e.g., parks, airports) where geometry is essential for reasoning. By systematically combining these geometric types (e.g., polyline–polygon, polyline–bounding box), EarthSpatialBench covers a rich spectrum of real-world spatial configurations and allows EarthSpatialBench to evaluate both object-level and region-level spatial reasoning.

\begin{itemize}[leftmargin=*,noitemsep,topsep=0pt]
\item \textbf{Bounding boxes.} Used for small, visually identifiable objects such as buildings and facilities. For example, \textit{“Locate the building (bounding box) closest to the railway (polyline).”}
\item \textbf{Polygons.} Represent areal regions such as parks, agricultural fields, and lakes. For example, \textit{“How many river segments (polyline) intersect this park (polygon)?”}
\item \textbf{Polylines.} Capture linear features such as roads, railways, and rivers. For example, \textit{“Does this road segment (polyline) intersect the river (polyline)?”}
\end{itemize}

\textbf{Object Representations.} 
Objects in Earth imagery can be referenced through multiple representation modalities. EarthSpatialBench supports three complementary object representation methods: textual descriptions, visual overlays, and explicit geometric coordinates.

\begin{itemize}[leftmargin=*,noitemsep,topsep=0pt]
    \item \textbf{Textual descriptions.} Objects are specified through semantic attributes and spatial context, such as object types, functional roles, and surrounding features. For example, \textit{unique object references} (e.g., the only hospital in the image), \textit{extreme position references} (e.g., the northernmost building), and \textit{spatial context references} (e.g., the building north of a unique parking lot). This format evaluates whether models can ground diverse linguistic cues in complex geospatial scenes.
    
    \item \textbf{Visual overlays.} Target objects are highlighted directly on the image using visual markers or masks. Visual overlays facilitate explicit visual grounding and reduce ambiguity in object referencing.
    
    \item \textbf{Geometric coordinates.} Objects are represented through precise spatial specifications in the form of bounding boxes, polylines, or polygons. This format enables geometry-aware reasoning and tests whether models can leverage structured spatial inputs.
\end{itemize}

\textbf{Question Formats.} Beyond the spatial relation and geometry involved, the formulation of a question determines the type of reasoning an MLLM must perform. EarthSpatialBench includes three complementary question formats: choice-based, quantitative, and localization, each probing a different aspect of spatial reasoning.

\begin{itemize}[leftmargin=*,noitemsep,topsep=0pt]
    \item \textbf{Choice-based questions.} These require selecting from a discrete set of options, such as Yes or No, or one of the cardinal directions. Choice-based formats emphasize categorical spatial decisions and are less sensitive to small numerical deviations, making them reliable for robust comparison across models.
    \item \textbf{Quantitative questions.} These require explicit numerical reasoning, including counting objects, estimating distances, or estimating angles. Quantitative tasks test whether MLLMs can integrate spatial scale, extent, and relative geometry rather than relying solely on visual recognition.
    \item \textbf{Localization questions.} These focus on spatial grounding by asking the model to identify or locate specific objects or regions that satisfy a spatial constraint. Localization tasks assess whether an MLLM can connect abstract relational reasoning back to concrete image regions, which is essential for practical geospatial analysis.
\end{itemize}

\begin{figure*}[ht]
    \centering
    \subfloat[]{%
        \includegraphics[height=0.25\textwidth]{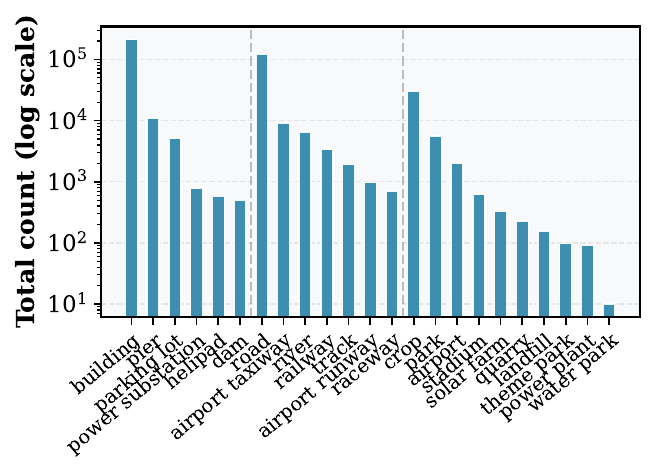}
    }
    \subfloat[]{%
        \includegraphics[height=0.25\textwidth]{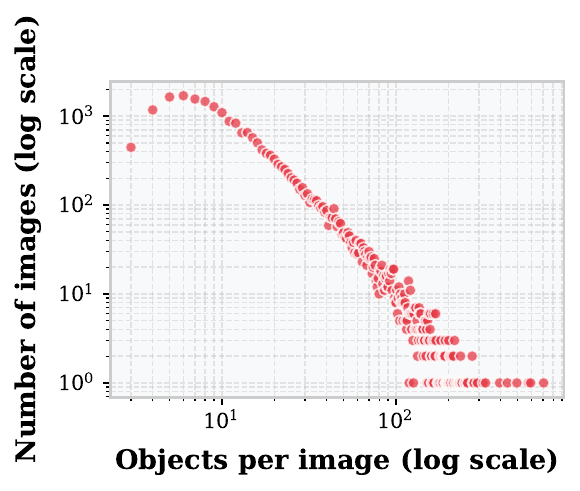}
    }
    \subfloat[]{%
        \includegraphics[height=0.28\textwidth]{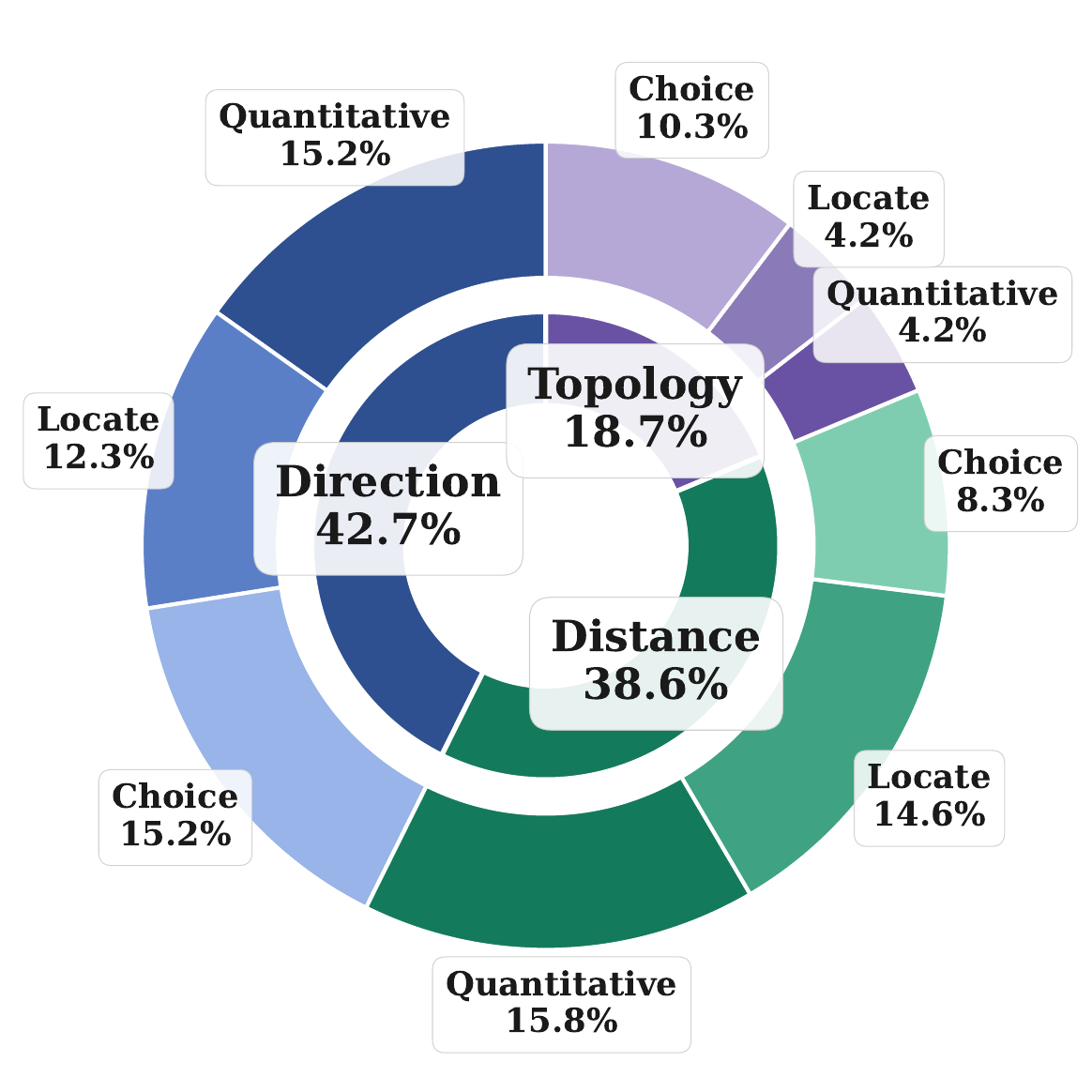}
    }
    \caption{(a) Distribution of all 23 geometric object types in EarthSpatialBench (log scale). The dataset includes 6 small-object bounding-box classes (building, dam, parking lot, pier, power substation, helipad), 7 polyline classes (airport runway, airport taxiway, raceway, railway, river, road, track), and 10 large-region polygon classes (airport, crop, landfill, park, quarry, solar farm, water park, theme park, stadium, power plant).
(b) Distribution of the total number of annotated objects per image on a log–log scale, showing the wide variation in scene complexity across images.
(c) Distribution of question counts across spatial relation categories (distance, direction, topology) and question formats (choice-based, quantitative, localization).}
\label{fig:distribution}
\end{figure*}

\subsection{Dataset Construction}
\textbf{Data Source.} We collect source data from SatlasPretrain~\cite{bastani2023satlaspretrain}, a large-scale publicly available remote sensing dataset containing high-resolution imagery paired with geometric object annotations. We first retain only images that contain geometric objects with visually discernible footprints, focusing primarily on buildings and other man-made structures. To ensure the presence of non-trivial spatial relationships, we further require that each image contains at least two distinct semantic categories beyond generic building footprints (e.g., river, road, park, dam). This avoids degenerate cases where only a single object type appears and no meaningful geometric relations can be formed.

\textbf{Geometry Processing and Spatial Relation Computation.} SatlasPretrain provides polygon and polyline labels for a variety of object classes. We preserve polygons for medium- and large-scale regions such as stadiums or airports. For small-scale objects such as buildings and parking lots whose polygon shapes may be too fine-grained for consistent visual interpretation, we convert their polygonal annotations into axis-aligned bounding boxes. For every pair of geometry instances in an image, we compute spatial relationships following standard GIS procedures. Distance is computed using Euclidean metrics for geometries (the nearest distance for each point in each object), and buffer queries use a consistent metric scale derived from georeferenced pixel resolution. Direction is computed using the azimuth angle between object centroids and discretized into 8-direction bins. Topology is computed using the DE-9IM model, the standard formalism for topological relations in computational geometry. We combine computed spatial relations with curated question templates for each reasoning category (distance, direction, topology). Templates cover choice-based, quantitative, and localization tasks. The final dataset includes multiple question types per relation, with both positive and negative exemplars to avoid answer biases. Because all answers are derived from exact geometry computations, EarthSpatialBench provides unambiguous ground-truth labels.

\textbf{Quality Control.} To ensure that EarthSpatialBench provides reliable spatial reasoning tasks, we implement a multi-stage quality assurance pipeline that combines automated checks with targeted manual review. We first perform automated checks to confirm that all polygons, polylines, and bounding boxes fall within image bounds, have non-zero area or length. Objects that are too small, noisy, or visually ambiguous (such as chimneys, wells, mineshafts, or flagpoles) are removed. We also exclude geometry pairs whose spatial relations are unstable or ambiguous, including cases where boundaries are nearly touching and produce inconsistent topological results under minor geometric variation. We then conduct manual review on a selected subset of samples that require closer inspection. These include (i) images containing very large geometric regions, (ii) scenes with high object density (for example, more than 100 buildings), (iii) objects intersecting image boundaries, and (iv) polygons with highly complex shapes or a large number of vertices. Observations from this review are used to refine question templates, remove problematic examples, and adjust geometric thresholds. To ensure comprehensive coverage, we also verify that each spatial relation type includes diverse examples across geometry types and object categories. Through this multi-stage verification process, EarthSpatialBench maintains high geometric fidelity, semantic clarity, and visual interpretability, enabling robust evaluation of spatial reasoning in Earth imagery.

\subsection{Dataset Statistics}

EarthSpatialBench is constructed to provide large-scale, geometry-rich supervision for evaluating spatial reasoning in multimodal models. Table~\ref{tab:dataset_overview} summarizes the key dataset properties. The benchmark contains 21K optical RGB images sourced from SatlasPretrain, each captured at approximately 1 m ground resolution and standardized to 512 × 512 pixels. On average, each image contains 19.3 annotated geometric entities, including 10.7 bounding-box objects, 6.7 polylines, and 1.9 polygonal regions, reflecting the heterogeneous composition of real-world Earth imagery.

Figure~\ref{fig:distribution} provides additional insights into dataset composition. The distribution of geometry types indicates substantial variation in the spatial structures present across scenes, ranging from dense built environments to large-region land-cover polygons. The wide spread in object counts per image demonstrates that the benchmark captures both simple and highly complex spatial layouts, which is essential for evaluating model robustness under varying scene densities. The distribution of question types further confirms that the benchmark spans a broad range of reasoning demands, ensuring that no single relation type or question format dominates the evaluation. These statistics show that EarthSpatialBench offers a large, diverse, and geometrically complex foundation for systematically evaluating MLLMs’ spatial reasoning capabilities in Earth imagery.

\begin{table*}[ht]\scriptsize
\centering
\caption{Performance on \textbf{Distance} reasoning tasks. }
\label{tab:distance}
{
\begin{tabular}{l|cc|cc|cc|cc}
\toprule
\multirow{2}{*}{Model} &
\multicolumn{2}{c|}{Exact Distance} &
\multicolumn{2}{c|}{Count} &
\multicolumn{2}{c|}{Localization} &
\multicolumn{2}{c}{Binary Classification} \\

& MAE $\downarrow$ & RMSE $\downarrow$
& MAE $\downarrow$ & RMSE $\downarrow$
& F1 $\uparrow$ & IoU $\uparrow$
& Acc $\uparrow$ & F1 $\uparrow$ \\
\midrule

Qwen3-VL-I-30B
& 127.93 & 168.31 & 1.28 & 2.06 & 0.25 & 0.18 & 0.58 & 0.56 \\

Qwen3-VL-I-30B-CoT
& 124.78 & 169.16 & 1.19 & \underline{1.41} & 0.16 & 0.16 & 0.53 & 0.34 \\

Qwen3-VL-T-30B
& 114.87 & 158.92 & \underline{1.02} & 1.77 & \underline{0.25} & \underline{0.19} & 0.71 & 0.48 \\

Qwen3-VL-T-30B-CoT
& 121.17 & 158.39 & 1.33 & 1.68 & \textbf{0.26} & \textbf{0.20} & 0.56 & 0.34 \\

InternVL3.5-38B
& 144.21 & 189.73 & \textbf{0.89} & \textbf{1.19} & 0.23 & 0.19 & 0.59 & 0.54 \\

InternVL3.5-38B-CoT
& 131.51 & 174.07 & 1.21 & 1.47 & 0.16 & 0.13 & 0.60 & 0.53 \\

Claude-3.5-Sonnet
& 146.51 & 191.15 & 1.47 & 2.21 & 0.12 & 0.05 & 0.60 & 0.59 \\

Claude-3.5-Sonnet-CoT
& 115.86 & 155.57 & 1.24 & 1.63 & 0.07 & 0.02 & 0.69 & 0.66 \\

Gemini 2.5 Pro
& 119.44 & 161.97 & 1.90 & 3.27 & 0.23 & 0.12 & \underline{0.74} & \underline{0.74} \\

Gemini 2.5 Pro-CoT
& \underline{97.82} & \underline{134.22} & 1.65 & 1.77 & 0.24 & 0.10 & 0.72 & 0.71 \\

GPT-5
& \textbf{93.91} & 150.61 & 1.73 & 2.85 & 0.18 & 0.07 & \textbf{0.86} & \textbf{0.86} \\

GPT-5-CoT
& 100.23 & \textbf{129.66} & 1.15 & 1.62 & 0.12 & 0.04 & 0.70 & 0.67 \\

\bottomrule
\end{tabular}
}
\end{table*}

\begin{table*}[ht]\scriptsize
\centering
\caption{Performance on \textbf{Direction} reasoning tasks. }
\label{tab:direction}
{
\begin{tabular}{l|cc|cc|cc|cc|cc}
\toprule
\multirow{2}{*}{Model}

& \multicolumn{2}{c|}{Exact Angle}
& \multicolumn{2}{c|}{Count}
& \multicolumn{2}{c|}{Localization}

& \multicolumn{2}{c|}{8-Direction Classification}
& \multicolumn{2}{c}{Binary Classification} \\

& MAE $\downarrow$ & RMSE $\downarrow$
& MAE $\downarrow$ & RMSE $\downarrow$
& F1 $\uparrow$ & IoU $\uparrow$

& Acc $\uparrow$ & F1 $\uparrow$
& Acc $\uparrow$ & F1 $\uparrow$ \\
\midrule

Qwen3-VL-I-30B        
& 117.20 & 145.86 & 2.88 & 6.37 & \textbf{0.24} & \textbf{0.16}
& 0.35 & 0.31 & 0.67 & 0.67 \\

Qwen3-VL-I-30B-CoT    
& 80.31  & 130.91 & 1.74 & 5.21 & 0.04 & 0.05
& 0.34 & 0.34 & 0.57 & 0.38 \\

Qwen3-VL-T-30B       
& 117.98 & 140.94 & 1.64 & \textbf{2.90} & \underline{0.21} & \underline{0.13}
& 0.26 & 0.26 & \underline{0.71} & \underline{0.71} \\

Qwen3-VL-T-30B-CoT   
& \textbf{54.99}  & \underline{96.28}  & 2.24 & 6.37 & 0.16 & 0.11
& 0.45 & 0.38 & 0.70 & 0.46 \\

InternVL3.5-38B         
& 126.31 & 158.25 & 1.55 & 3.14 & 0.16 & 0.11
& 0.30 & 0.29 & 0.64 & 0.64 \\

InternVL3.5-38B-CoT     
& 71.06  & 113.33 & \underline{1.47} & 3.30 & 0.16 & 0.10
& 0.25 & 0.24 & 0.54 & 0.50 \\

Claude-3.5-Sonnet       
& 130.18 & 154.41 & 1.61 & 3.10 & 0.07 & 0.03
& 0.36 & 0.31 & 0.67 & 0.67 \\

Claude-3.5-Sonnet-CoT   
& 65.68  & 112.24 & 1.74 & 3.32 & 0.10 & 0.03
& 0.45 & 0.41 & 0.70 & 0.68 \\

Gemini 2.5 Pro           
& 134.47 & 159.88 & 2.02 & 3.91 & 0.09 & 0.03
& 0.37 & 0.31 & 0.67 & 0.67 \\

Gemini 2.5 Pro-CoT       
& 61.20  & 97.58  & 1.69 & 3.71 & 0.07 & 0.03
& \underline{0.48} & \underline{0.46} & 0.66 & 0.62 \\

GPT-5                   
& 119.26 & 142.71 & 1.76 & 3.34 & 0.12 & 0.05
& 0.43 & 0.41 & 0.70 & 0.70 \\

GPT-5-CoT               
& \underline{55.66}  & \textbf{95.78}  & \textbf{1.46} & \underline{2.97} & 0.11 & 0.04
& \textbf{0.51} & \textbf{0.49} & \textbf{0.73} & \textbf{0.72} \\

\bottomrule
\end{tabular}
}
\end{table*}

\begin{table*}[ht]\scriptsize
\centering
\caption{Performance on \textbf{Topological} reasoning tasks.}
\label{tab:topology}
{
\begin{tabular}{l|cccc|cccc|cccc}
\toprule
\multirow{2}{*}{Model}

& \multicolumn{4}{c|}{Count}
& \multicolumn{4}{c|}{Localization}
& \multicolumn{4}{c}{Binary Classification} \\

& \multicolumn{2}{c}{Intersect}
& \multicolumn{2}{c|}{Within}

& \multicolumn{2}{c}{Intersect}
& \multicolumn{2}{c|}{Within}

& \multicolumn{2}{c}{Intersect}
& \multicolumn{2}{c}{Within} \\

& MAE $\downarrow$ & RMSE $\downarrow$
& MAE $\downarrow$ & RMSE $\downarrow$

& F1 $\uparrow$ & IoU $\uparrow$
& F1 $\uparrow$ & IoU $\uparrow$

& Acc $\uparrow$ & F1 $\uparrow$
& Acc $\uparrow$ & F1 $\uparrow$ \\
\midrule

Qwen3-VL-I-30B        
& 1.14 & \textbf{1.82} & \textbf{2.45} & \textbf{4.30}
& 0.24 & \underline{0.19} & 0.13 & 0.15
& 0.68 & 0.63 & 0.90 & 0.71 \\

Qwen3-VL-I-30B-CoT    
& 1.10 & 1.98 & 2.83 & 6.55
& 0.19 & 0.17 & 0.15 & 0.13
& 0.71 & 0.69 & 0.94 & 0.77 \\

Qwen3-VL-T-30B       
& 1.07 & \underline{1.88} & 2.79 & 6.54
& 0.21 & 0.18 & \textbf{0.21} & \underline{0.20}
& 0.73 & 0.72 & 0.91 & 0.68 \\

Qwen3-VL-T-30B-CoT   
& 1.07 & 2.09 & 3.06 & 6.84
& 0.19 & 0.18 & 0.12 & 0.07
& 0.72 & 0.72 & 0.94 & 0.68 \\

InternVL3.5-38B         
& 1.17 & 2.00 & 3.23 & 6.86
& 0.15 & 0.13 & 0.04 & 0.09
& 0.60 & 0.58 & 0.92 & 0.65 \\

InternVL3.5-38B-CoT     
& 1.26 & 2.07 & 3.27 & 6.87
& 0.16 & \underline{0.19} & 0.02 & \textbf{0.42}
& 0.72 & 0.70 & 0.93 & 0.59 \\

Claude-3.5-Sonnet       
& 1.23 & 1.97 & 2.83 & 6.49
& 0.14 & 0.15 & 0.13 & 0.06
& 0.61 & 0.56 & 0.89 & 0.68 \\

Claude-3.5-Sonnet-CoT   
& \underline{1.06} & 1.91 & 3.25 & 7.09
& 0.22 & \underline{0.19} & 0.12 & 0.07
& 0.73 & 0.71 & 0.95 & 0.70 \\

Gemini 2.5 Pro           
& \textbf{1.05} & 1.93 & 2.94 & 6.90
& 0.12 & 0.06 & 0.01 & 0.00
& 0.77 & 0.77 & \textbf{0.98} & \underline{0.91} \\

Gemini 2.5 Pro-CoT       
& 1.09 & 2.01 & 3.02 & 7.05
& \underline{0.25} & 0.18 & 0.01 & 0.00
& \underline{0.80} & \underline{0.79} & \underline{0.97} & \textbf{0.93} \\

GPT-5                   
& 1.34 & 2.09 & \underline{2.60} & \underline{6.25}
& 0.09 & 0.12 & \textbf{0.21} & 0.12
& 0.75 & 0.73 & 0.96 & 0.85 \\

GPT-5-CoT               
& 1.24 & 2.05 & 3.17 & 6.91
& \textbf{0.26} & \textbf{0.27} & \underline{0.19} & 0.11
& \textbf{0.82} & \textbf{0.81} & \underline{0.97} & \underline{0.88} \\

\bottomrule
\end{tabular}
}
\end{table*}

\section{Experiment}

\subsection{Experimental Setup}

\textbf{Evaluated MLLMs.} We evaluated a representative set of both open-source and proprietary  multimodal LLMs. The open-source models include Qwen3-VL-30B-Instruct and Qwen3-VL-30B-Thinking~\citep{Qwen3-VL2025} and InternVL3.5-38B~\citep{wang2025internvl3}. The proprietary models include GPT-5, Gemini 2.5 Pro, and Claude 3.5 Sonnet.

\textbf{Evaluation Protocol.} 
We adopt task-specific metrics aligned with the three question formats in EarthSpatialBench. For quantitative questions (e.g., counting, distance and angle estimation), we report mean absolute error (MAE) and root mean squared error (RMSE) with respect to the ground truth. For choice-based questions (e.g., Yes/No or directional classification), we report overall accuracy and macro-averaged F1 score. For localization questions, which require identifying 2D bounding boxes satisfying spatial constraints, we evaluate intersection-over-union (IoU) and macro F1 score based on the overlap between predicted and ground-truth boxes, using an IoU threshold of 0.1. 

\textbf{Computational Resources.} 
All open-source models are evaluated on a compute node equipped with 8 NVIDIA B200 GPUs. Proprietary models are accessed through their respective APIs using default inference configurations.

\subsection{Main Results}
Tables~\ref{tab:distance}, \ref{tab:direction}, and \ref{tab:topology} summarize the performance of multiple MLLMs with chain-of-thought (CoT) prompting on distance, direction, and topological reasoning tasks. Overall, we observe substantial performance gaps across reasoning types, indicating that current models exhibit uneven capabilities in numerical estimation, spatial grounding, and geometric reasoning.

For quantitative tasks, GPT-5 achieves the lowest error in distance estimation, while Qwen3-VL-Thinking and its CoT variant demonstrate strong performance on angle regression, suggesting superior numerical reasoning ability. For localization tasks, Qwen3-VL variants generally outperform other models, whereas proprietary models exhibit comparatively weaker performance on distance- and direction-related localization. For classification tasks, large proprietary models achieve the best performance, reflecting strong symbolic and relational reasoning capabilities. CoT prompting consistently improves classification and regression performance for most models, particularly on topological reasoning tasks, but often yields limited or negative gains for localization, suggesting that explicit reasoning traces do not directly translate to improved spatial grounding.

Finally, strong reasoning performance does not necessarily imply robust visual grounding. For example, Claude and Gemini attain competitive classification accuracy but perform poorly on localization metrics, revealing a clear decoupling between high-level spatial reasoning and low-level visual grounding. These results indicate that, despite recent advances, multimodal LLMs remain limited in jointly reasoning over precise geometry, numerical relations, and visual evidence in Earth observation scenarios.

\subsection{Grounding as a Prerequisite for Spatial Reasoning}

\begin{figure}[ht]
    \centering
    \includegraphics[width=0.45\textwidth]{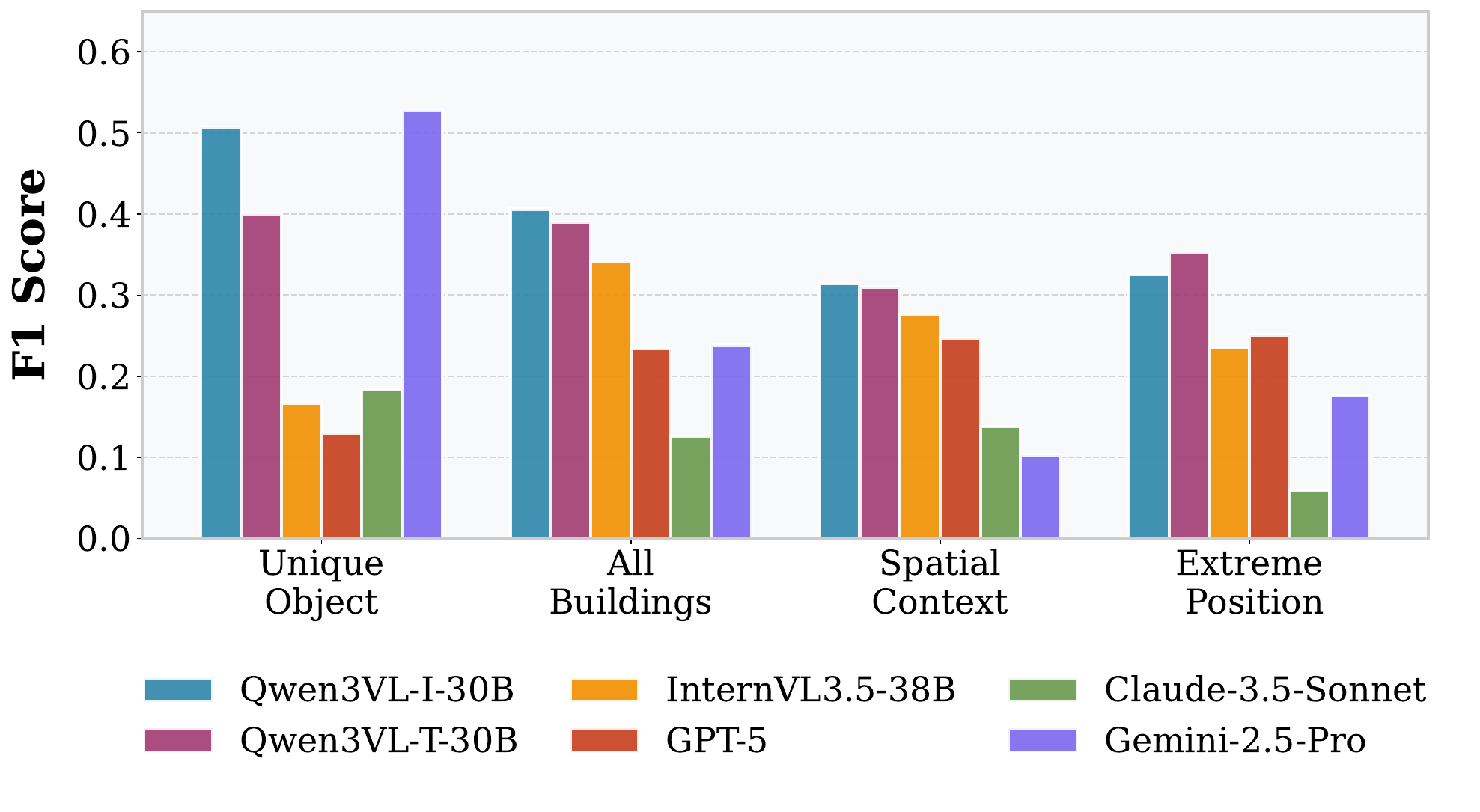}
    \caption{Performance of MLLMs on four grounding variants.}
\label{fig:grounding}
\end{figure}

Most spatial reasoning tasks depend on accurate object localization, as geometric relations computed using GIS procedures require precise spatial inputs. We therefore evaluate grounding as a prerequisite for reliable spatial reasoning. Figure~\ref{fig:grounding} reports F1 scores on four grounding variants that isolate object identification and localization difficulty. We observe substantial model-dependent variation, with Qwen3-VL variants achieving the strongest and most consistent performance, while GPT-5 and Claude exhibit marked degradation on exhaustive grounding (\textit{all buildings}). Gemini 2.5 Pro performs well on \textit{unique object} grounding but degrades on context-dependent and extreme cases, indicating limited generalization to ambiguous scenarios. Across models, \textit{extreme position} objects remains particularly challenging, highlighting rare-attribute grounding as a major bottleneck. Overall, these results demonstrate that strong spatial reasoning is tightly coupled with robust low-level grounding, and that joint grounding and reasoning in complex Earth imagery remains an open challenge.

\subsection{Comparisons on Different Geometric Representations}

\begin{figure*}[ht]
    \centering
    \subfloat[]{%
        \includegraphics[height=0.23\textwidth]{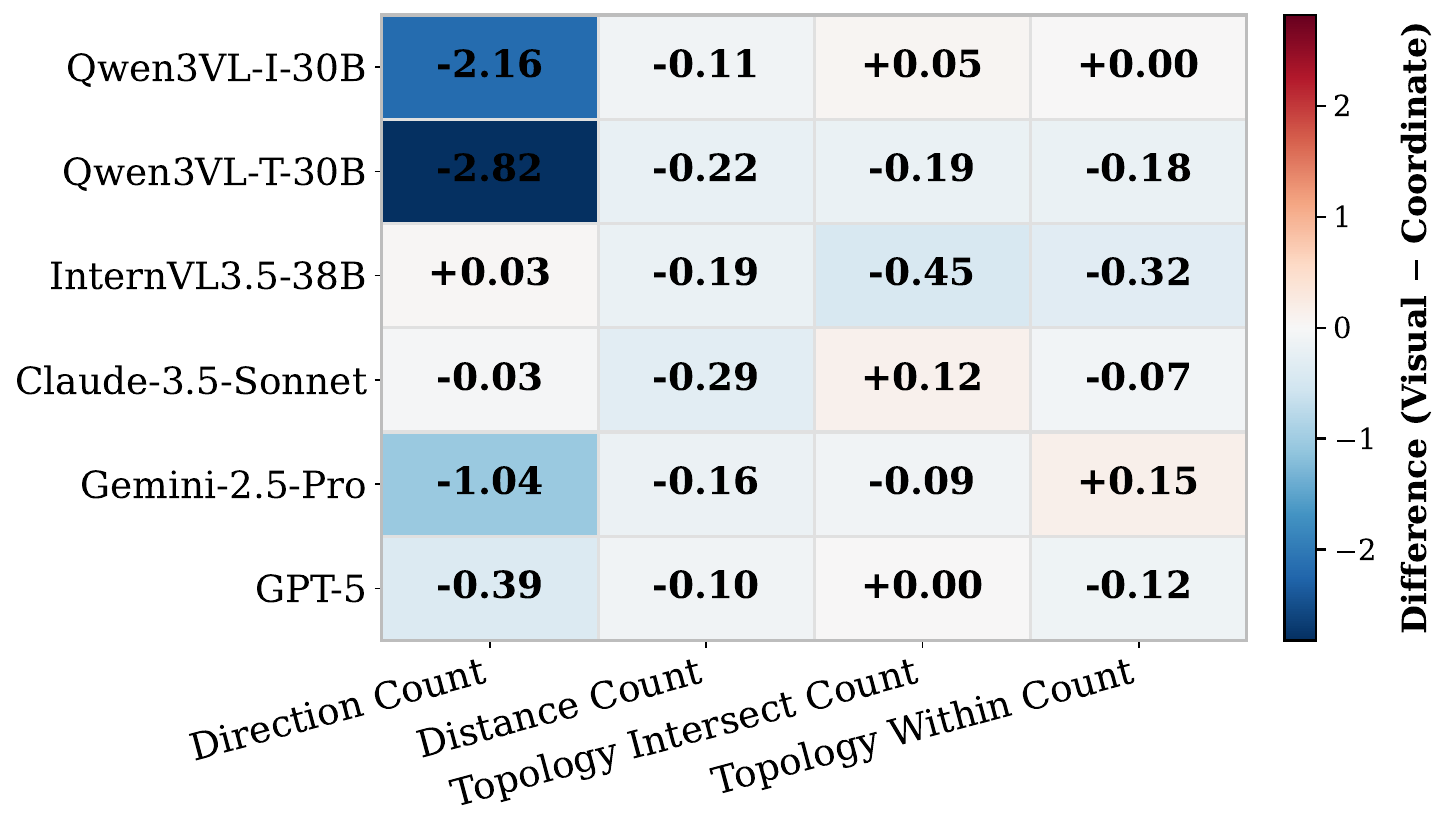}
    }
    \subfloat[]{%
        \includegraphics[height=0.23\textwidth]{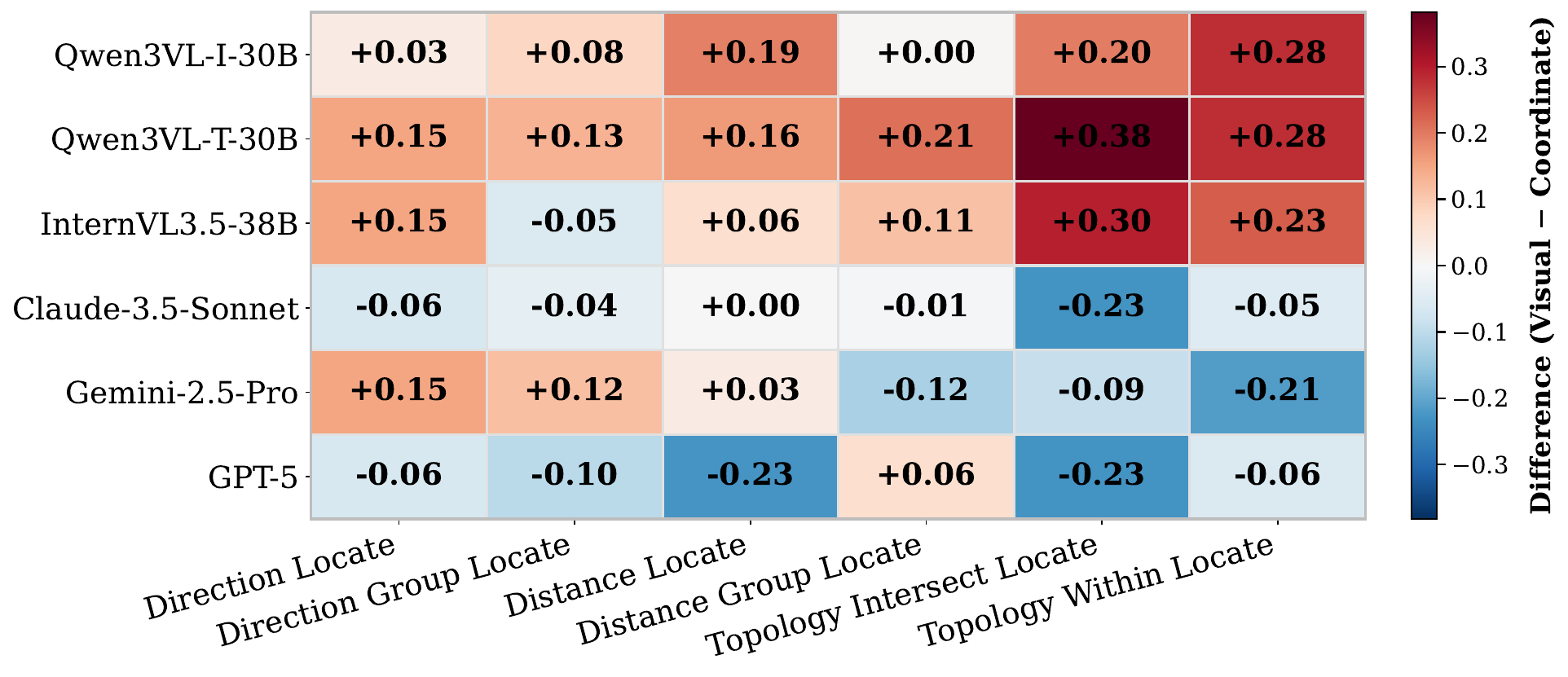}
    }
    \caption{Performance differences between visual-overlay-based and coordinate-based object references. (a) Change in MAE for counting. (b) Change in F1 score for localization. }
\label{fig:heatmap}
\end{figure*}

Figure~\ref{fig:heatmap} compares model performance when spatial references are provided as visual overlays versus explicit coordinates. Overall, visual overlays improve localization performance, while their impact on counting and regression tasks is more heterogeneous. Across models, we observe a clear divergence between grounding-oriented and reasoning-oriented behaviors. Qwen3-VL models consistently leverage visual overlays to improve localization and counting, demonstrating strong vision–language alignment. In contrast, GPT-5 achieves competitive performance under coordinate-based descriptions but fails to effectively exploit visual overlays, revealing a preference for text-centric reasoning. These results highlight substantial differences in how current MLLMs internalize spatial information and suggest that future systems must better unify visual grounding and geometric reasoning for robust Earth observation analysis.

\begin{figure}
    \centering
    \includegraphics[width=0.45\textwidth]{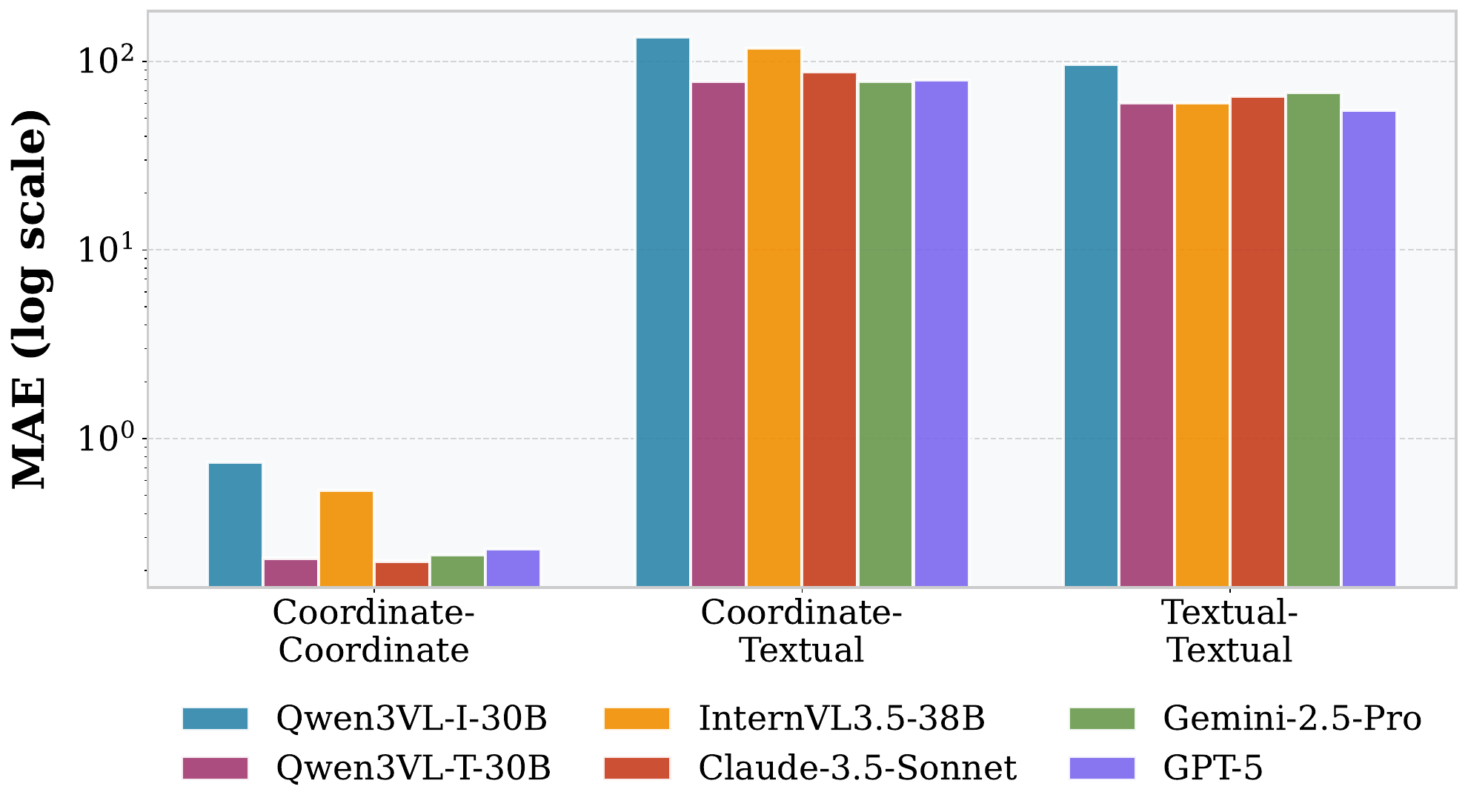}
    \caption{MAE of angle estimation under different object representation settings.}
    \label{fig:placeholder}
\end{figure}

Figure~\ref{fig:placeholder} further examines the impact of object representation on direction estimation by comparing coordinate-based and language-based references. When explicit coordinates are provided for both objects (\textit{coordinate–coordinate}), all models achieve substantially lower MAE, indicating that they can reliably perform angle computation through direct numerical reasoning. In contrast, performance degrades markedly when objects are specified using unique textual descriptions, highlighting the difficulty of grounding linguistic references to precise spatial locations. Notably, the \textit{coordinate–textual} setting yields even higher errors than \textit{textual–textual}, suggesting that mixing symbolic coordinates with ambiguous language descriptions may introduce additional grounding inconsistencies. These results emphasize that accurate spatial reasoning depends critically on reliable object grounding, and that partial access to geometric information is insufficient when linguistic grounding remains uncertain.

\subsection{Comparisons on Different Geometric Types}

\begin{figure}[ht]
    \centering
    \includegraphics[width=0.45\textwidth]{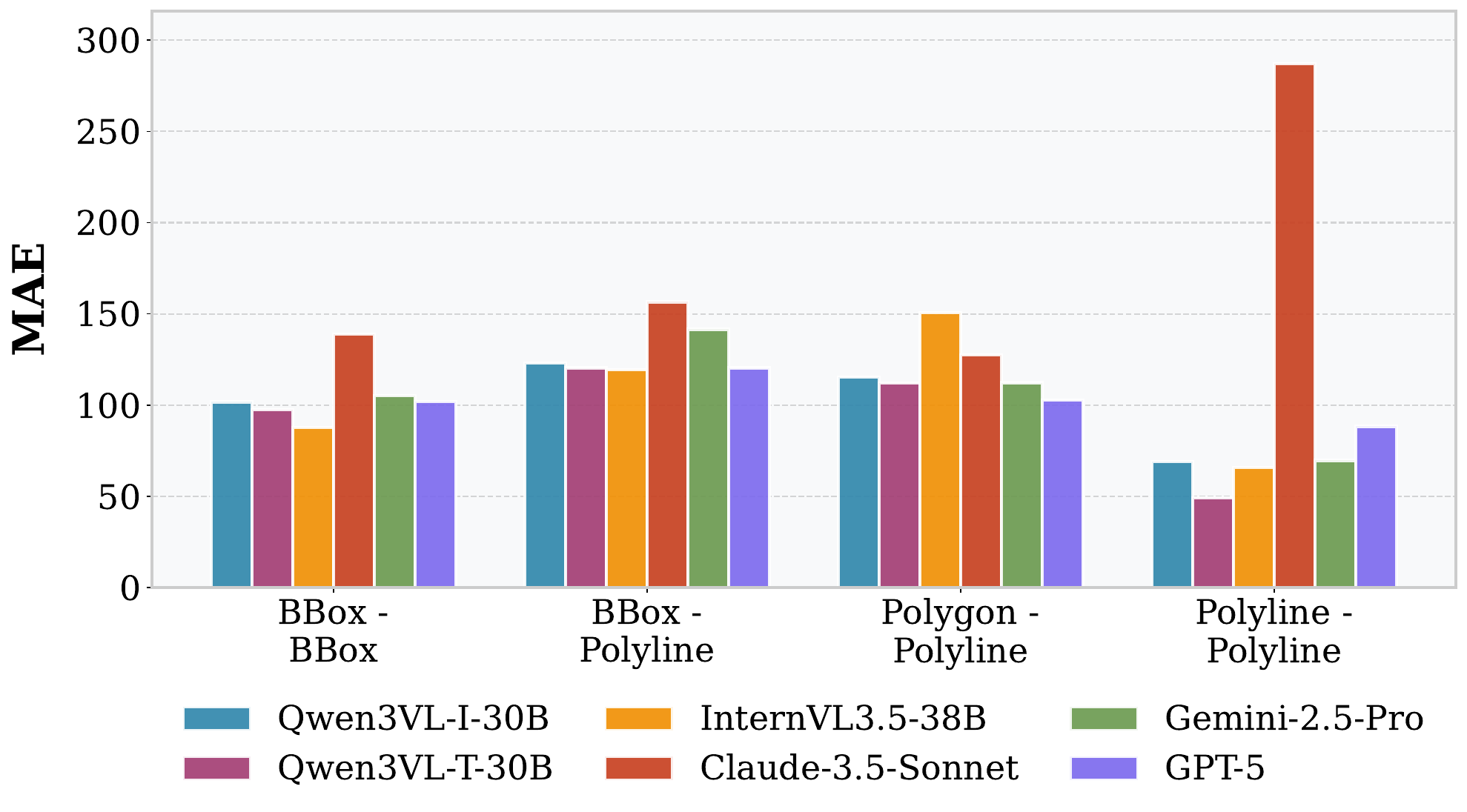}
    \caption{MAE of distance estimation across different geometry type combinations.}
\label{fig:geometry_distance}
\end{figure}

Figure~\ref{fig:geometry_distance} reports the MAE of distance estimation across different geometry type combinations. We observe substantial performance variations induced purely by geometric representation, with heterogeneous pairs (BBox--Polyline and Polygon--Polyline) consistently yielding higher errors than homogeneous cases. Notably, polyline--polyline queries achieve lower errors than box--box queries for most models, suggesting that linear structures such as roads and coastlines provide clearer geometric cues than coarse bounding boxes. In contrast, polygon--polyline settings remain among the most challenging, reflecting difficulties in reasoning over irregular region–curve interactions. From a model perspective, Qwen3-VL variants exhibit relatively stable performance across geometry types, whereas GPT-5 and Claude show pronounced degradation on polyline-based queries. These results highlight geometric representation as a critical yet underexplored factor and motivate the incorporation of explicit geometric reasoning mechanisms in future models.

\section{Conclusion and Future Work}
We introduced EarthSpatialBench, a comprehensive benchmark for evaluating spatial reasoning in multimodal large language models on Earth imagery. By formalizing spatial relations into distance, direction, and topology and integrating diverse geometric entities and question formats, EarthSpatialBench enables the systematic assessment of spatial intelligence in real-world geospatial settings. Our experiments reveal limitations of state-of-the-art MLLMs across all reasoning categories, highlighting the gap between current vision–language capabilities and the demands of practical geospatial analysis. Looking forward, extending the benchmark to multi-temporal and multi-modal inputs (e.g., DEM, SAR, and LiDAR) represents promising directions for future research.




\section*{Impact Statement}
This work introduces EarthSpatialBench, a benchmark for evaluating spatial reasoning capabilities of multimodal large language models on Earth imagery. By facilitating systematic evaluation of spatial understanding in remote sensing and aerial imagery, this benchmark has the potential to support socially beneficial applications such as disaster response, urban planning, environmental monitoring, and infrastructure assessment.

Overall, we believe that this work primarily advances methodological research in multimodal machine learning and geospatial intelligence, and we do not anticipate significant negative societal impacts when used responsibly.


\bibliography{main}
\bibliographystyle{icml2026}

\newpage
\appendix
\onecolumn


\end{document}